\documentclass{article} 
\usepackage{iclr2022_conference,times}

\usepackage{amsmath,amsfonts,bm}

\def\eqref#1{equation~\ref{#1}}

\def\1{\bm{1}}

\DeclareMathAlphabet{\mathsfit}{\encodingdefault}{\sfdefault}{m}{sl}
\SetMathAlphabet{\mathsfit}{bold}{\encodingdefault}{\sfdefault}{bx}{n}

\usepackage{graphicx}
\usepackage{hyperref}
\usepackage{url}
\usepackage{enumitem}

\title{KDSTM: Neural Semi-supervised Topic Modeling with Knowledge Distillation}

\author{Weijie Xu\textsuperscript{1}, Xiaoyu Jiang\textsuperscript{1}, Jay Desai\textsuperscript{1}, Bin Han\textsuperscript{2}, Fuqin Yan\textsuperscript{1} \& Francis Iannacci\textsuperscript{1} \\
\textsuperscript{1} Amazon Inc. \textsuperscript{2} University of Washington\\
\texttt{\{weijiexu,billyjia,jdesa,fqinyan,iannacci\}@amazon.com} \\
\texttt{\{binhan96816\}@gmail.com} \\
}

\iclrfinalcopy 
\begin{document}

\maketitle

\begin{abstract}
In text classification tasks, fine tuning pretrained language models like BERT and GPT-3 yields competitive accuracy; however, both methods require pretraining on large text datasets. In contrast, general topic modeling methods possess the advantage of analyzing documents to extract meaningful patterns of words without the need of pretraining. 
To leverage topic modeling's unsupervised insights extraction on text classification tasks, we develop the Knowledge Distillation Semi-supervised Topic Modeling (KDSTM). KDSTM requires no pretrained embeddings, few labeled documents and is efficient to train, making it ideal under resource constrained settings. Across a variety of datasets, our method outperforms existing supervised topic modeling methods in classification accuracy, robustness and efficiency and achieves similar performance compare to state of the art weakly supervised text classification methods.

\end{abstract}

\section{Introduction}
The current state-of-the-art language modeling methods often require transfer learning \cite{brown2020language}, large amount of labels \cite{yang2019xlnet} and pretrained embeddings \cite{cao2020multilingual}. Consequently, they are difficult to apply in low resource settings, where many endangered languages\cite{austin2011cambridge} lack both pre trained language models and sufficient labeled documents. 
For semi-supervised method \cite{meng2018weaklysupervised} tailored to limited label scenario, it is time consuming to both tune and train. 

Topic modeling is an unsupervised method for discovering latent structure within the training document sets and achieves great empirical performance in many fields\cite{blei2009nested}, including finance \cite{https://doi.org/10.1111/eufm.12326}, healthcare \cite{DBLP:journals/corr/abs-1711-10960}, education \cite{zhao2020targeted}, marketing \cite{Reisenbichler2019} and social science \cite{762586}. \cite{jelodar2018latent} provides a survey on the applications of topic modeling. 

Latent Dirichlet Allocation (LDA) \cite{blei2003latent} is the most fundamental topic modeling approach based on Bayesian inference on Markov chain Monte Carlo (MCMC) and variational inference; however, it is hard to be expressive or capture large vocabularies. Neural topic model (NTM) \cite{miao2018discovering} leverages auto-encoding \cite{kingma2014semi} framework to approximate intractable distributions over latent variables. Recently, embedded topic model (ETM) \cite{dieng2020topic} uses word embedding during the reconstruction process to make topic more coherent and reduce the influence of stop words. The goal of unsupervised topic modeling methods \cite{blei2003latent,teh2006hierarchical,miao2018discovering,gemp2019weakly, xu-etal-2023-vontss} is to maximize the probability of the observed data, resulting in the tendency to identify obvious and superficial aspects of a corpus.  To incorporate users' domain knowledge of documents into the model, supervised modeling \cite{blei2010supervised, JMLR:v13:zhu12a, pmlr-v108-wang20c} has been studied. However, supervised methods do not perform well when the labeled set is small.

In this work, we propose knowledge distillation semi-supervised topic modeling (KDSTM), which only requires a few labeled documents for each topic as input. KDSTM  utilizes knowledge distillation and optimal transport to guide topic extraction with seed documents. 
It achieves state of the art results when benchmarking with supervised topic modeling and weakly supervised text classification methods. 

Advantages of KDSTM are summarized as follows:
\begin{itemize}[leftmargin=0.12in]
    \item KDSTM is a novel architecture which incorporates knowledge distillation and optimal transport into the neural topic modeling framework.
    \item KDSTM consistently achieves better topics classification performance on different datasets when compared to supervised topic modeling methods or weakly supervised text classification methods. 
    \item KDSTM only requires a limited number of labeled documents as input, making it more practical in low resource settings.
    \item KDSTM does not rely on any transfer learning or pre trained language models. The embedding is trained on the dataset, making it suitable for less common/endangered languages. 
    \item KDSTM is efficient to train and fine-tune compared to existing methods. This makes it suitable to be trained and run inference on resource constrained devices.

\end{itemize}

\section{Preliminary}
\textbf{Supervised Topic Modeling}
Since topic modeling reduces the dimensionality of the text, the learned low dimensional topic distributions can be used in the downstream tasks. \cite{blei2010supervised} adds a response variable associated with each document and assumes that the variable can be fitted by Gaussian distribution to make it tractable. Labeled LDA \cite{ramage2009labeled} assumes that each document can be associated with one topic and uses this information to create the model. Recently, BP-SLDA \cite{chen2015endtoend} uses back propagation to make LDA supervised. Dieng \cite{dieng2017topicrnn} incorporates RNN with LDA to make latent variables more suitable for downstream tasks. TAM \cite{wang2020neural} combines GSM \cite{liu2019neural} and RNN \cite{2020} to do the supervised topic modeling. To be specific, it uses GSM to fit a document generative process and estimates document specific topic distribution. It uses GRU \cite{chung2014empirical} to encoded word tokens. After that, it jointly optimizes two components by using an attention mechanisms. Recently, topic modeling is also combined with Siamese network \cite{huang2018siamese} to achieve better prediction performance. However, these methods' performances drops when training set is small.

\textbf{Optimal Transport}
To avoid matching labels with multiple topics 
, we consider the optimal transport distance \cite{chen2019improving, torres2021survey}, which has been widely used for comparing the distribution of probabilities. Specifically, let $U(r,c)$ be the set of positive $m \times n$ matrices for which the rows sum to r and the columns sum to c:
$U(r, c) = \{P \in R_{>0}^{m \times n}|P 1_{t} = r, P^{T} 1_{s} =c \}$ For each position t, s in the matrix, it comes with a cost $M_{t,s}$. Our goal is to solve $d_{M}(r, c) = min_{P \in U(r, c)} \sum_{t, s} P_{t,s} M_{t,s} $. To make distribution homogeneous \cite{cuturi2013sinkhorn}, we let \begin{equation} d_{M}^{\lambda}(r, c) = min_{P \in U(r, c)} \sum_{t,s} P_{t,s} M_{t, s} - \frac{1}{\lambda} h(P)\end{equation}, where $h(P) = - \sum_{t,s} P_{t,s} \log P_{t,s}$. Optimal Transport induces good robustness and semantic invariance in NLP related tasks \cite{chen2019improving} or topic modeling \cite{zhao2020neural, xu2018distilled}.

\textbf{Knowledge Distillation}
Knowledge distillation is the process of transferring knowledge from a large model to a smaller one. Given a large model trained for a specific classification task, the final layer of the network is a softmax in the form: \begin{equation}y(x|\tau) = \frac{e^{\frac{s(x)}{\tau}}}{\sum_{j}e^{\frac{s(x)}{\tau}} }\end{equation}
where $t$ is the temperature parameter. The softmax operator converts the logit values $s$ to pseudo-probabilities. Knowledge distillation consists of training a smaller network, called the distilled model, on a separate dataset called the transfer set. Cross entropy is used as the loss function between the output of the distilled model and output produced by the large model on the transfer set, using a high value of softmax temperature $\tau$ for both models. \cite{hinton2015distilling}
\begin{equation} E(\mathbf {x} |\tau)=-\sum _{i}{\hat {y}}_{i}(\mathbf {x} |\tau)\log y_{i}(\mathbf {x} |\tau) \end{equation}
where ${\hat {y}}_{i}$ is generated by the large model and $y_{i}$ is generated by the distilled model. Instead of using the prediction itself, few methods leverage similarity \cite{tung2019similaritypreserving, Chen2018DarkRankAD, passalis2020probabilistic} or features \cite{romero2015fitnets, passban2020alpkd, chen2021crosslayer} as guidance.
\begin{figure*}
\includegraphics[scale = 0.265]{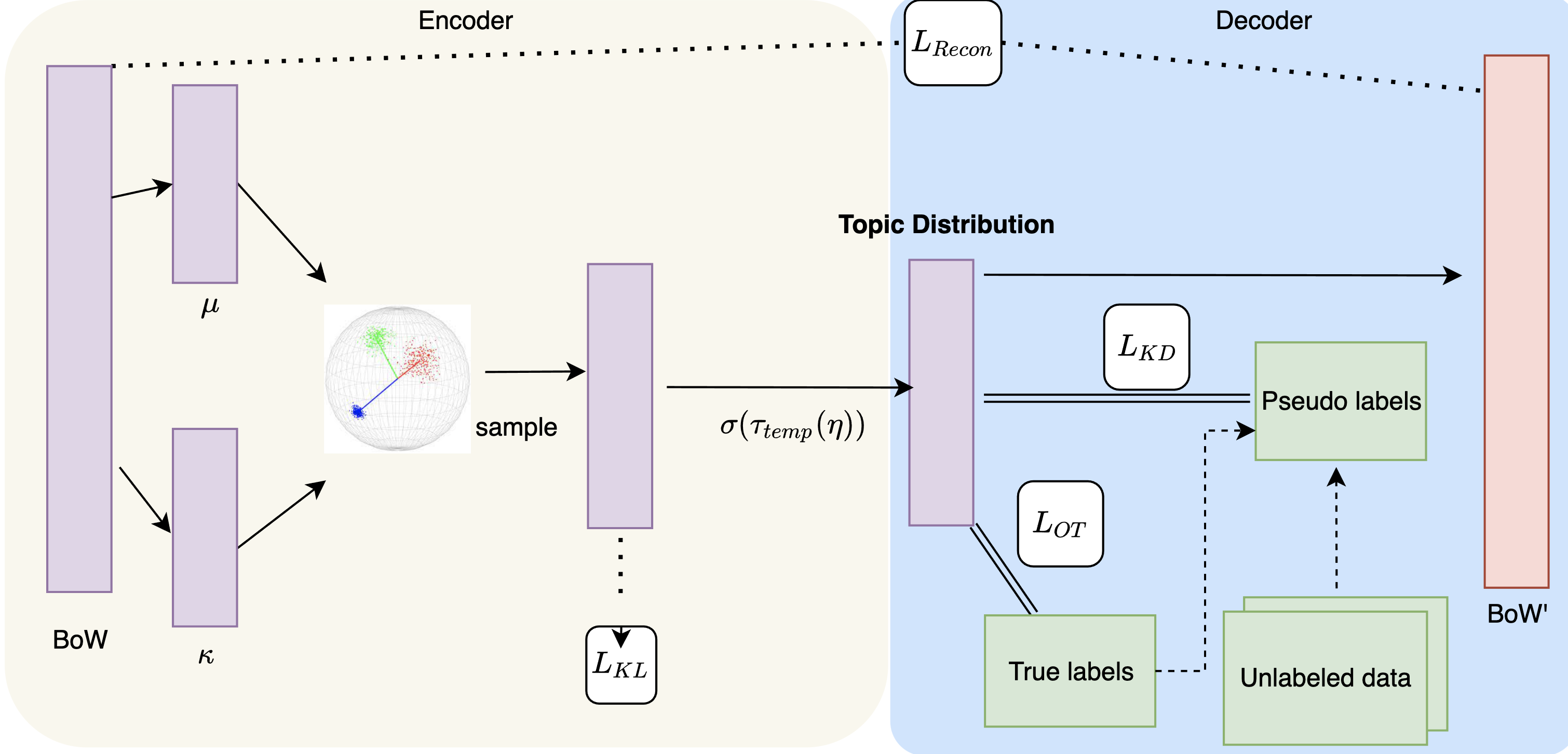}

\caption{The Architecture of KDSTM with four main loss function including reconstruction loss, KL divergence, optimal transport loss and knowledge distillation loss}
\label{fig:people4}
\end{figure*}

\section{Method}
The encoder network $\phi$ encodes the bag of words representation of any document $x_{d}$ and output parameters for latent distribution which can be used to sample the topic distribution $t_{d}$. Following \cite{dieng2020topic}, the decoder is represented by a vocabulary embedding matrix $e_{W}$ and a topic embedding matrix $e_{T}$. We use spherical word embedding \cite{meng2019spherical} to create $e_{W}$. We train it on the corpus and keep it fixed during the training. $W$ is the corpus and $T$ contains all topics. We also use VMF distribution (Appendix~\ref{vmf}) instead of normal distribution as the latent distribution for better clusterability \cite{xu2018spherical,davidson2018hyperspherical, reisinger2010spherical, batmanghelich2016nonparametric, 9547420}. In this notation, our modified ETM's algorithm can be described as follows: for every document $d$, 1) Generate  $t_{d}$ using sampled direction parameter $\mu$ and scale parameter $\kappa$ from $\phi$. 2) Reconstruct bag of words by $t_{d} \times softmax(e_{T} e_{W}^{T})$. The goal of ETM is to maximize the marginal likelihood of the documents: $\sum_{d = 1}^{D} \log p(x_{d} | e_{T}, e_{W})$. To make it tractable, the loss function combines reconstruction loss with KL divergence. The description of notation can be found in Table~\ref{sample-table} in appendix.  

In KDSTM, we adopt optimal transport to assign topics to labels. Each entry in M is defined as $M_{g,t} = 1 - mean_{x \in g}\phi_{t}(x)$ where $g$ is one of the labeled documents' group. Let $M_{g,t}$ represent the weights of words in labeled documents in group $g$ on topic $t$. We use sinkhorn distance as loss function and give high entropy $\lambda$ to make sure that each labeled comment falls into separate topics. Thus, \begin{equation} L_{OT} =  min_{P \in U(|T|, |G|)} \sum_{t,g} P_{t,g} M_{t, g} - \frac{1}{\lambda} h(P) \label{eq2}
\end{equation} where $|G|$ is the number of labeled groups and g represents one group of labeled comments.\\

The next step is to ensure that a test input text that is similar to our labeled document has high probability of being classified as the same topic. To achieve that, we borrow the idea from similarity and feature based knowledge distillation \cite{mun2018learning, tung2019similaritypreserving}. To be specific, we first train the unsupervised topic modeling till convergence. Then, we store the direction parameters $\mu$ from $\phi$ and use it to calculate cosine similarity $s$ between unlabeled and labeled documents. We use the maximum similarity in each labeled group as guidance for knowledge distillation. The benefits of this approach include: 1) latent distribution from unsupervised topic model can be used for label classification \cite{10.5555/2981780.2981892, chen2015endtoend}. These distributions can serve as teachers. 2) we do not need a separate and larger model, making it more resource efficient.\\
For each text $x_{d}$, we find the most similar document i in each labeled group $g$ and their similarity $s_{i}(x)$. Then we define \begin{equation} \hat {y^{g}}(x; \tau) = \frac{e^{\frac{\max_{i \in g} s_{i}(x)}{\tau}}}{\sum_{g \in G}e^{\frac{\max_{i \in g} s_{i}(x)}{\tau}}}\label{eq5}
\end{equation} where G is all labeled text groups. The knowledge distillation loss is measured by : \begin{equation} L_{KD} = -\tau^{2} \sum_{d \in D} \sum_{g \in G} I(s_{g}(x_{d}) \geq thresh) \hat {y^{g}}(x_{d}; \tau) log(\phi(x_{d}))\label{eq6}
\end{equation} We use the indicator function $I(s_{i}(x_{d}) \geq thresh)$. Since we only have few labels available, some of documents may not be related to any of the labeled documents. Thus, we only care about documents that are relevant to existing labels to provide a better teaching experience. We split training into 3 stages: 1) we train the standard topic modeling with KL Divergence and reconstruction loss $L_{KD} + L_{Recon}$ till convergence. We calculate similarity matrix $s$ after this stage. 2) We add optimal transport loss $L_{KD} + L_{Recon} + \alpha L_{OT}$ and train for few epochs. 3) We add knowledge distillation loss $L_{KD} + L_{Recon} + \alpha L_{OT} + \beta L_{KD}$ and train for few epochs. In practice, step 2 and step 3 are less time consuming and thus, this helps user optimize their labels in online settings.
The architecture is illustrated in Figure~\ref{fig:people4}. \cite{hoyle2020improving} also leverages knowledge distillation, but uses bag of words representation, is of unsupervised nature, and using a BERT-based auto-encoder as the teacher.

\section{Experiments}
\textbf{Settings} In this section, we report the experimental results for our methods and three additional state of the arts methods (WestClass, LLDA, TAM). To form the vocabulary, we keep all words that appear more than a certain number of times and vary the threshold from 20 to 100 depending on the size of the dataset. We remove documents that are less than 2 words. We also remove stop words, digits, time and symbols from vocabulary and use a fully-connected neural network with two hidden layers of [256, 64] unit and ReLU as the activation function followed by a dropout layer(rate = 0.5). The hyperparameter setting used for all baseline models and vNTM are similar to \cite{JMLR:v20:18-569}. We use Adam\cite{kingma2017adam} as the optimizer with learning rate 0.002 and use batch size 256. We use \cite{smith2018superconvergence} as scheduler and use learning rate 0.01 for maximally iterations equal to 50. For each run, we sample 5 documents per class and use them as inputs, calculate performance metrics on the rest of unlabeled documents, run each algorithm 10 times and report the result in the bin plot. We report accuracy, aucroc and averaged micro f1 score. For hyperparameters, we use $\lambda = 50$, $\alpha = \beta = 10$, $thresh = 0$ and $\tau = 1$ for our method and perform moderate tuning on parameters presented in the original papers of other methods.  Our code is written in PyTorch and all the models are trained on AWS using ml.p2.8xlarge (NVIDIA K80).

\begin{figure*}
\includegraphics[scale = 0.34]{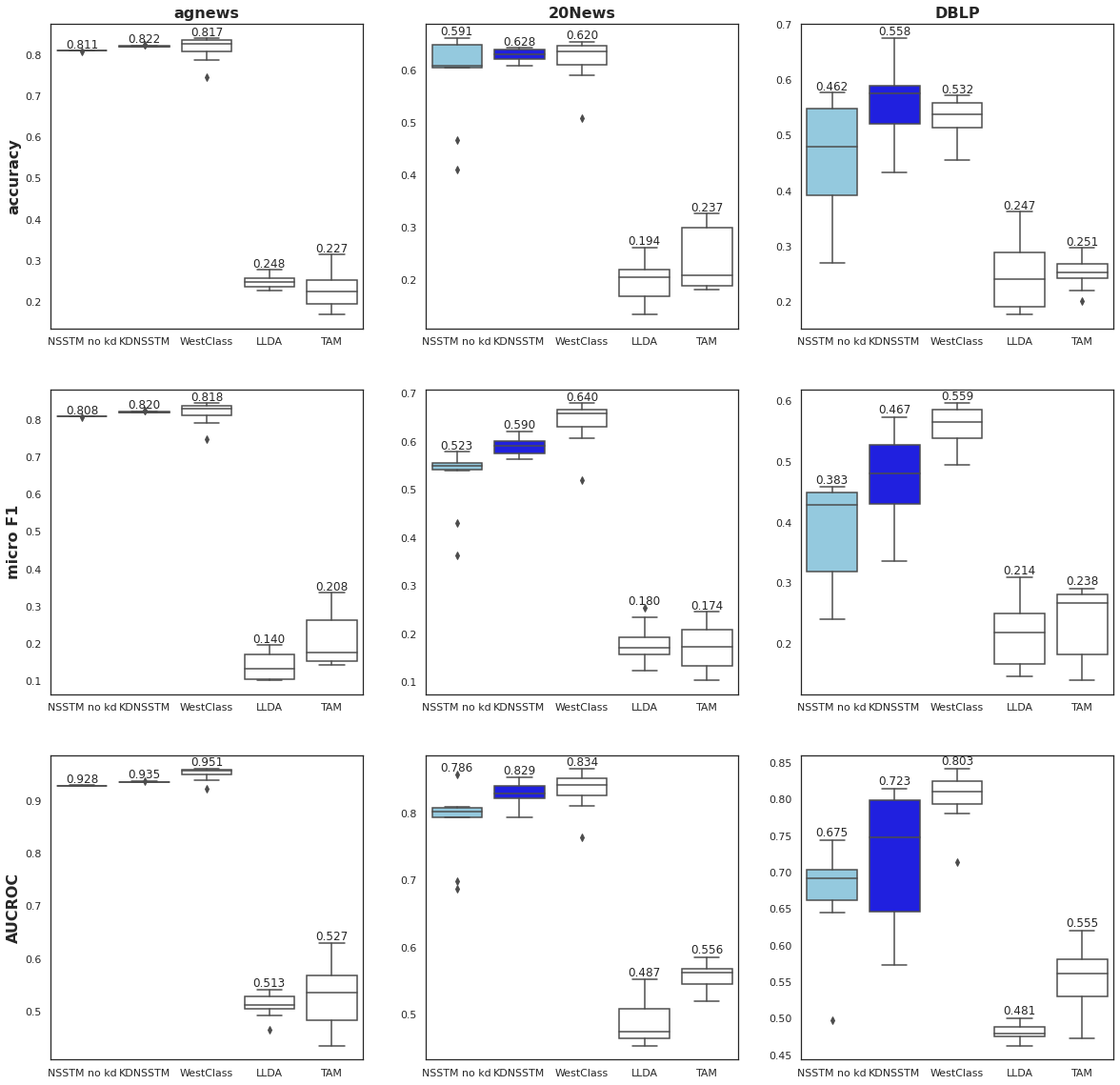}

\caption{From left to right, is NSSTM without knowledge distillation (Sky Blue),  KDNSSTM (Blue), WestClass, LabeledLDA, TAM. We average result as text on top of each bin plot }
\label{fig:people3}
\end{figure*}

\textbf{Datasets} (1) \textbf{AG's News} We use AG's News dataset from \cite{zhang2016characterlevel}. It has 4 classes and 30000 documents per class. Class categories include World, Sports, Business and Sci/Tech for evaluation; (2) \textbf{DBLP} \cite{Tang:08KDD, Tang:10TKDD} dataset consists of bibliography data in computer science. DBLP selects a list of conferences from 4 research areas, database, data mining, artificial intelligence, and computer vision. With a total 60,744 papers averaging 5.4 words in each title, DBLP tests the performance on small text corpus. See Appendix~\ref{dataset}. (3) \textbf{20News} \cite{lang1995newsweeder} is a collection of newsgroup posts. We only select 4 categories here. Compared to the other two datasets, 4 categories newsgroup is small so that we can check the performance of our methods on small datasets.

\textbf{Methods} (1) \textbf{WestClass} \cite{meng2018weaklysupervised}: This method is a weakly-supervised neural text classification method. The weak supervision source can come from any of the three sources: label surface names, class related keywords, and labeling documents. Output will be document labels consistent with cluster of inputs. We will use this method to benchmark the performance on document classification accuracy.  (2) \textbf{L-Label} \cite{ramage2009labeled}: This method accepts labeled documents and is used to compare KDSTM. To better benchmark the performance, we fine-tune alpha, eta and number of iterations to get the best accuracy performance. (3) \textbf{TAM} \cite{pmlr-v108-wang20c}: This method achieves the better performance than existing supervised topic modeling methods. To benchmark the performance for a strongly performing model, we fine-tune the dimension of GRU gate, learning rate and number of epochs. See Appendix~\ref{code}.

\textbf{Results} As can be seen from Figure~\ref{fig:people3}, KDSTM consistently performs significantly better than existing supervised topic modeling methods on all 3 classification metrics (accuracy, F1, and AUC). On standard dataset like AG's News, our performance is higher than WestClass on accuracy and micro F1. Despite similar overall performance, our method is on average 4 times faster than WestClass (Table~\ref{table:6} in appendix). If we finetune $\tau$, our method can further improve which show its potential to beat WestClass (Table~\ref{fig:scaoe} in appendix). Our method has high variance in DBLP where each document has on average 5.4 words. We also compare the performance without knowledge distillation. Knowledge distillation increases the performance of the model on all 3 metrics. Knowledge distillation provides less benefit for small dataset such as 20News.

\section{Conclusion}
In this work, we develop Knowledge Distillation Semi-supervised Topic Modeling (KDSTM) using knowledge distillation and optimal transport. Our method achieves improved performance across several classification metrics compared to existing supervised topic modeling methods and more efficient than existing weakly supervised text classification methods. Our method does not require transfer learning or pretrained embeddings and is faster to train and fine-tune, making it ideal in low resource scenarios. For future work, we will extend this work to include sequential information to further improve its performance and stability.

\bibliography{iclr2022_conference}
\bibliographystyle{iclr2022_conference}
\newpage
\appendix
\textbf{\Large Appendix}

\section{Notations Tables}
Table~ref{sample-table} summarizes the notation used in the paper.
\begin{table}[t]
\caption{Description of the notations used in this work.}
\label{sample-table}

\vskip 0.15in

\begin{center}
\begin{small}
\begin{tabular}{ll}
\hline
Notion & Description \\
\hline
$W$ & Corpus   \\
$M$ & cost matrix for topic modeling   \\
$s$ & similarity matrix between labeled documents and unlabeled documents  \\
$Z$ & topic proportions  \\
$t_{d}$ & topic distribution  \\
$X$ & bow of words representation for all documents \\
$x_{d}$ & bag of words representation for a document\\
$\phi$ & encoder   \\
$L_{Recon}(X)$ & reconstruction loss  \\
$e_{W}$ & word embedding matrix \\
$e_{T}$ & topic embedding matrix \\
$\mu$ & vmf direction parameter    \\
$\kappa$ &vmf concentration parameter  \\
$\tau$ &temperature parameter for knowledge distillation  \\
$P$ & wight matrix for optimal transport \\
$g$ & a labeled document group \\
$G$ & all labeled document group  \\
$T$ & group of all topics \\
$D$ & all documents \\
$L_{OT}$ & optimal transport loss  \\
$\lambda$ & entropy penalty weights  \\
$L_{KL}$ & KL divergence \\
$i$ & a labeled document from a group g \\
$s_{i}(x)$ & the similarity of document x on labeled document i \\
$L_{KD}$ & knowledge distillation loss \\
$\alpha, \beta$ & coefficients for  $L_{KD} $ and $ L_{OT}$\\
$thresh$ & threshold for knowledge distillation \\ 
\hline
\end{tabular}
\end{small}
\end{center}
\vskip -0.1in
\end{table}

\section{Time Tables}
We document the training time on our machine for all these methods and averaged across 10 runs. We also capture finetune time of KDSTM. To be specific, finetune time is stage 2 and 3 of KDSTM. Before this stage, model does not require any labels to train. 

\begin{table*}

\centering
\begin{tabular}{| c| c | c| c } 
\hline
\multicolumn{1}{c}{} \vline &
\multicolumn{1}{c}{20News}\vline  & \multicolumn{1}{c}{AG's News} \vline & \multicolumn{1}{c}{DBLP}\\
\hline
\multicolumn{1}{c}{TAM} \vline & 98.98 &  842.64 &  378.05\\
\multicolumn{1}{c}{KDSTM} \vline & 20.75 &  252.87 &  117.36\\
\multicolumn{1}{c}{KDSTM finetune} \vline & 9.04 & 106.01 & 43.72 \\
\multicolumn{1}{c}{WestClass} \vline & 104.30 & 888.61 & 87.17\\
\hline
\end{tabular}
\caption{We capture the training time(seconds) of methods WestClass, KDSTM overall/finetune stage and TAM.}
\label{table:6}
\end{table*}

\section{von Mises-Fisher Distribution} 

\label{vmf}
In low dimensions, the Gaussian density presents a concentrated probability mass around the origin. This is problematic when the data is partitioned into multiple clusters. An ideal prior should be non informative and uniform over the parameter space. Thus, the von Mises-Fisher(vMF) is used in VAE. vMF is a distribution on the (M-1)-dimensional sphere in $R^{M}$, parameterized by $\mu \in R^{M}$ where $||\mu|| = 1$ and a concentration parameter $\kappa \in R_{\geq 0}$. The probability density function of the vMF distribution for $z \in R^{D}$ is defined as:
$$q(Z|\mu, \kappa) = C_{M}(\kappa) exp(\kappa\mu^{T}Z)$$
$$C_{M}(\kappa) = \frac{\kappa^{\frac{M}{2} - 1}}{(2\pi)^{\frac{M}{2}} I_{\frac{M}{2} - 1}(\kappa)} + log 2$$
where $I_{v}$ denotes the modified Bessel function of the first kind at order v. The KL divergence with vMF(., 0) \cite{davidson2018hyperspherical} is
$$KL(vMF(\mu, \kappa)|vMF(.,0)) = \kappa\frac{I_{\frac{M}{2}}(\kappa)}{I_{\frac{M}{2}-1}(\kappa)} $$
$$+ (\frac{M}{2} - 1) log \kappa - \frac{M}{2} log (2\pi)  - log I_{\frac{M}{2}-1}(\kappa) $$ $$+ \frac{M}{2} log \pi + log2 + log \Gamma(\frac{M}{2})$$
vMF based VAE has better clusterability of data points especially in low dimensions \cite{guu2018generating}.

\section{Dataset Details} 
\label{dataset}
\begin{table*}
\centering
\begin{tabular}{llr}
\hline
Corpus Name & Class name (Number of documents in the class) & Average Document Length \\
\hline

20News & Atheism (689), Religion (521), Graphics (836), Space (856) & 55.6 \\
AG's News & \vtop{\hbox{\strut Politics (30000), Sports (30000),}\hbox{\strut Business (30000), Technology (30000)}} & 45.0 \\
DBLP & \vtop{\hbox{\strut Database (11981), Data Mining (4763), }\hbox{\strut Artificial Intelligence (20890), Computer Vision (16961)}} & 5.4 \\
\hline
\end{tabular}
\caption{Details of selected datasets
}
\label{table:4}
\end{table*}

\section{Code for Benchmarks}
\label{code}
\textbf{WestClass} \cite{meng2018weaklysupervised}: This method is a weakly-supervised neural text classification method. The weak supervision source can come from any of the three sources: label surface names, class related keywords, and labeling documents. Output will be document labels consistent with cluster of inputs. We will use this method to benchmark the performance on document classification accuracy. The code we use is: \url{https://github.com/yumeng5/WeSTClass}\\
\textbf{L-Label} \cite{ramage2009labeled}: This method accepts labeled documents and is used to compare KDSTM. To better benchmark the performance, we fine-tune alpha, eta and number of iterations to get the best accuracy performance. The code we use is: \url{https://github.com/JoeZJH/Labeled-LDA-Python}\\
\textbf{TAM} \cite{pmlr-v108-wang20c}: This method achieves the best performance compare to other supervised topic modeling methods. To better benchmark the performance, we fine-tune dimension of GRU gate, learning rate and num of epochs to get the best performance. Since this method is not stable, we show the median of 10 runs to get the baselines. The code we use is: \scalebox{0.8}{\url{https://github.com/WANGXinyiLinda/Neural-Topic-Model-with-Attention-for-Supervised-Learning}}

\section{Scale VS Performance}
\begin{figure*}
\includegraphics[scale = 0.25]{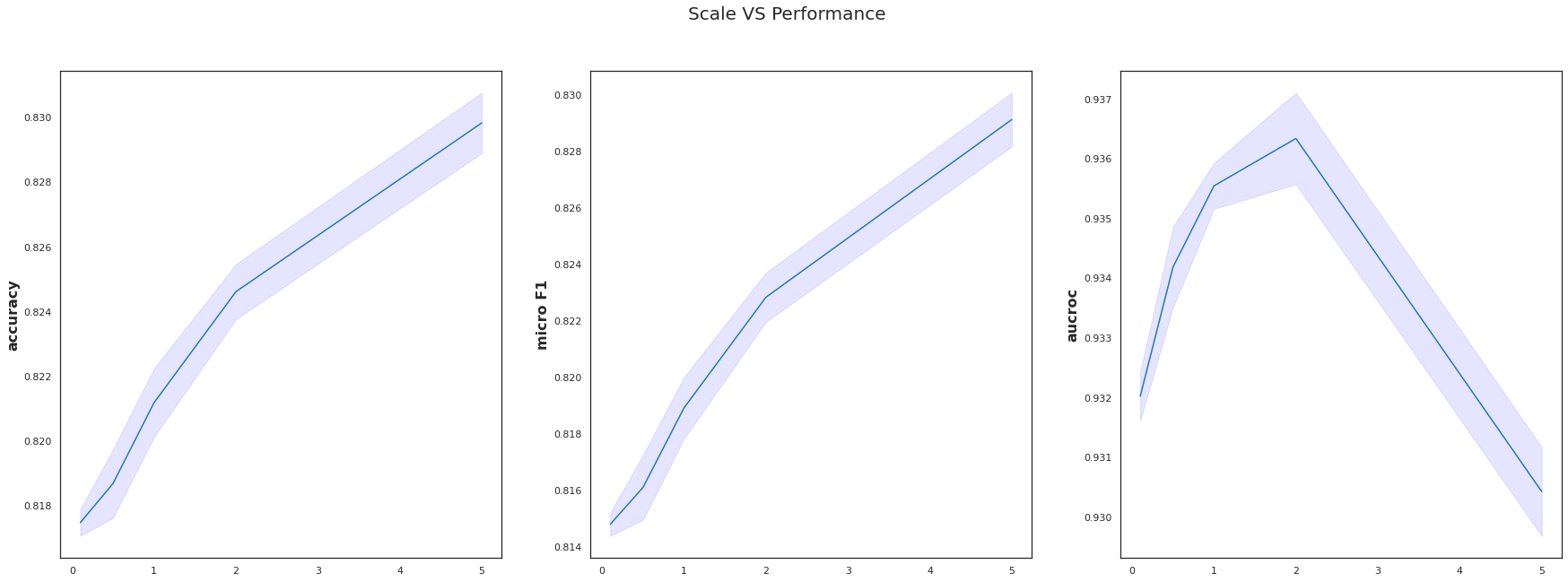}

\caption{Performance vs Scale}
\label{fig:scaoe}
\end{figure*}

Figure~\ref{fig:scaoe} shows the variation of metrics wrt scale. As scale increase, micro F1 and accuracy increase while AUC decreases.

\end{document}